\documentclass{article} 
\usepackage{iclr2017_conference,times}
\usepackage{hyperref}
\usepackage{url}
\usepackage{graphicx}
\usepackage{subfigure}
\usepackage{amssymb,amsmath,bm}

\title{Tacotron: Towards End-to-End Speech Synthesis}

\author{Yuxuan Wang\thanks{ These authors really like tacos.}~~,~RJ Skerry-Ryan$^\ast$,~Daisy Stanton, Yonghui Wu,~Ron J. Weiss\thanks{ These authors would prefer sushi.}~~,~Navdeep Jaitly, \And
Zongheng Yang,~Ying Xiao$^\ast$,~Zhifeng Chen,~Samy Bengio$^\dagger$,~Quoc Le,~Yannis Agiomyrgiannakis,\And
Rob Clark,~Rif A. Saurous$^\ast$
\\
\\
Google, Inc.\\
\texttt{\{yxwang,rjryan,rif\}@google.com}
}

\begin{document}

\maketitle

\begin{abstract}
A text-to-speech synthesis system typically consists of multiple stages, such as a text analysis frontend, an acoustic model and an audio synthesis module. Building these components often requires extensive domain expertise and may contain brittle design choices. In this paper, we present Tacotron, an end-to-end generative text-to-speech model that synthesizes speech directly from characters. Given $<$text, audio$>$ pairs, the model can be trained completely from scratch with random initialization. We present several key techniques to make the sequence-to-sequence framework perform well for this challenging task. Tacotron achieves a 3.82 subjective 5-scale mean opinion score on US English, outperforming a production parametric system in terms of naturalness. In addition, since Tacotron generates speech at the frame level, it's substantially faster than sample-level autoregressive methods.
\end{abstract}

\section{Introduction}
Modern text-to-speech (TTS) pipelines are complex \citep{taylor2009text}. For example, it is common for statistical parametric TTS to have a text frontend extracting various linguistic features, a duration model, an acoustic feature prediction model and a complex signal-processing-based vocoder \citep{zen2009statistical,agiomyrgiannakis2015vocaine}. These components are based on extensive domain expertise and are laborious to design. They are also trained independently, so errors from each component may compound. The complexity of modern TTS designs thus leads to substantial engineering efforts when building a new system.

There are thus many advantages of an integrated end-to-end TTS system that can be trained on $<$text, audio$>$ pairs with minimal human annotation. First, such a system alleviates the need for laborious feature engineering, which may involve heuristics and brittle design choices. Second, it more easily allows for rich conditioning on various attributes, such as speaker or language, or high-level features like sentiment. This is because conditioning can occur at the very beginning of the model rather than only on certain components. Similarly, adaptation to new data might also be easier. Finally, a single model is likely to be more robust than a multi-stage model where each component's errors can compound. These advantages imply that an end-to-end model could allow us to train on huge amounts of rich, expressive yet often noisy data found in the real world.

TTS is a large-scale inverse problem: a highly compressed source (text) is ``decompressed" into audio. Since the same text can correspond to different pronunciations or speaking styles, this is a particularly difficult learning task for an end-to-end model: it must cope with large variations at the signal level for a given input. Moreover, unlike end-to-end speech recognition \citep{chan2016listen} or machine translation \citep{wu2016google}, TTS outputs are continuous, and output sequences are usually much longer than those of the input. These attributes cause prediction errors to accumulate quickly. In this paper, we propose Tacotron, an end-to-end generative TTS model based on the sequence-to-sequence (seq2seq) \citep{sutskever2014sequence} with attention paradigm \citep{bahdanau2014neural}. Our model takes characters as input and outputs raw spectrogram, using several techniques to improve the capability of a vanilla seq2seq model. Given $<$text, audio$>$ pairs, Tacotron can be trained completely from scratch with random initialization. It does not require phoneme-level alignment, so it can easily scale to using large amounts of acoustic data with transcripts. With a simple waveform synthesis technique, Tacotron produces a 3.82 mean opinion score (MOS) on an US English eval set, outperforming a production parametric system in terms of naturalness\footnote{Sound demos can be found at \url{https://google.github.io/tacotron}}.

\begin{figure*}[t]
\centering
\includegraphics[scale=0.65]{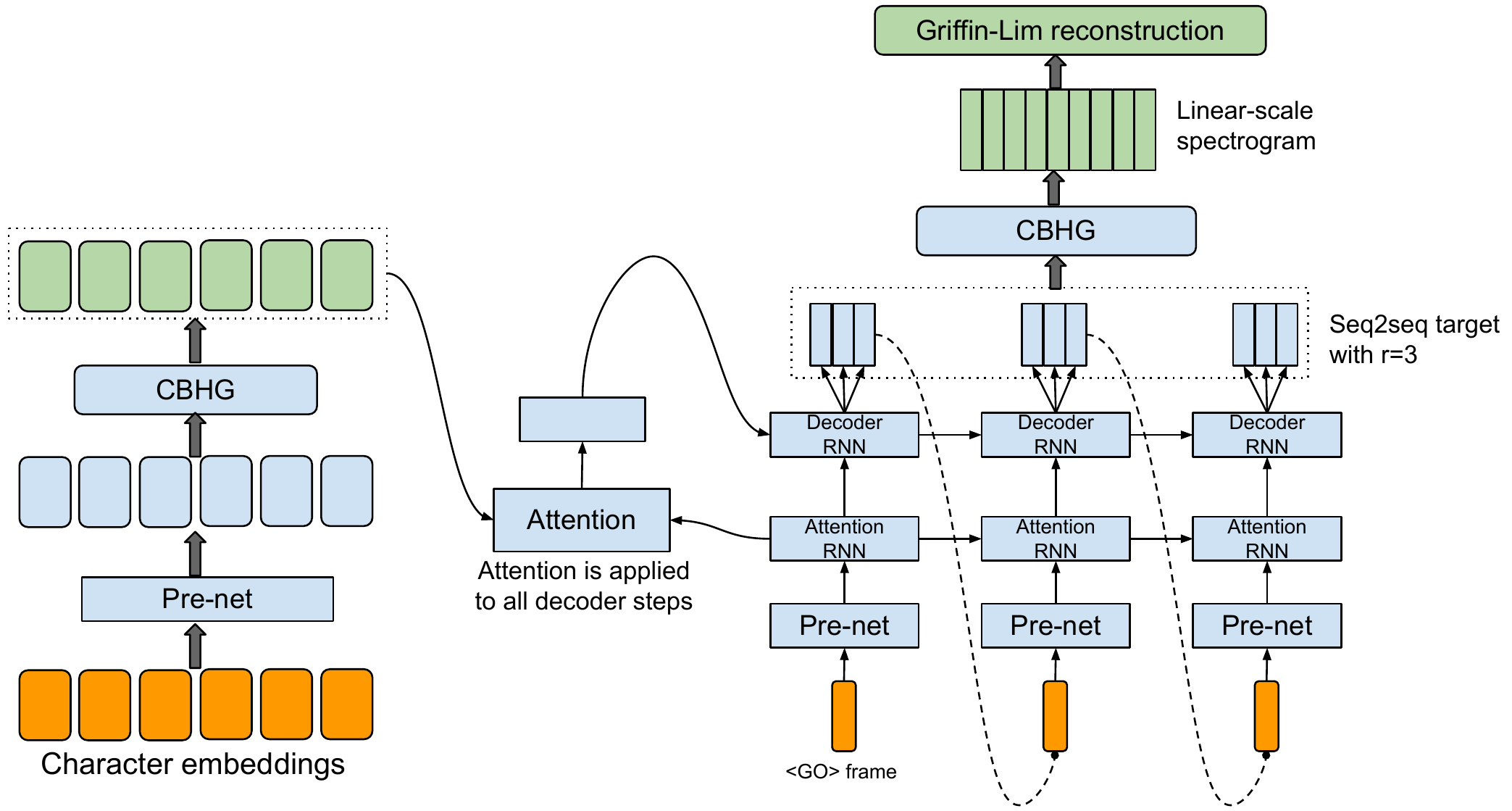}
\caption{{\it Model architecture. The model takes characters as input and outputs the corresponding raw spectrogram, which is then fed to the Griffin-Lim reconstruction algorithm to synthesize speech.}}
\label{fig:model}
\end{figure*}

\section{Related Work}
WaveNet \citep{van2016wavenet} is a powerful generative model of audio. It works well for TTS, but is slow due to its sample-level autoregressive nature. It also requires conditioning on linguistic features from an existing TTS frontend, and thus is not end-to-end: it only replaces the vocoder and acoustic model. Another recently-developed neural model is DeepVoice \citep{arik2017deep}, which replaces every component in a typical TTS pipeline by a corresponding neural network. However, each component is independently trained, and it's nontrivial to change the system to train in an end-to-end fashion.

To our knowledge, \cite{wang2016first} is the earliest work touching end-to-end TTS using seq2seq with attention. However, it requires a pre-trained hidden Markov model (HMM) aligner to help the seq2seq model learn the alignment. It's hard to tell how much alignment is learned by the seq2seq per se. Second, a few tricks are used to get the model trained, which the authors note hurts prosody. Third, it predicts vocoder parameters hence needs a vocoder. Furthermore, the model is trained on phoneme inputs and the experimental results seem to be somewhat limited.

Char2Wav \citep{sotelo2017char2wav} is an independently-developed end-to-end model that can be trained on characters. However, Char2Wav still predicts vocoder parameters before using a SampleRNN neural vocoder \citep{mehri2016samplernn}, whereas Tacotron directly predicts raw spectrogram. Also, their seq2seq and SampleRNN models need to be separately pre-trained, but our model can be trained from scratch. Finally, we made several key modifications to the vanilla seq2seq paradigm. As shown later, a vanilla seq2seq model does not work well for character-level inputs.

\section{Model Architecture}
\label{sec.model}
The backbone of Tacotron is a seq2seq model with attention \citep{bahdanau2014neural, vinyals2015grammar}. Figure \ref{fig:model} depicts the model, which includes an encoder, an attention-based decoder, and a post-processing net. At a high-level, our model takes characters as input and produces spectrogram frames, which are then converted to waveforms. We describe these components below.

\begin{figure}[htp!]
\centering
\includegraphics[scale=0.75]{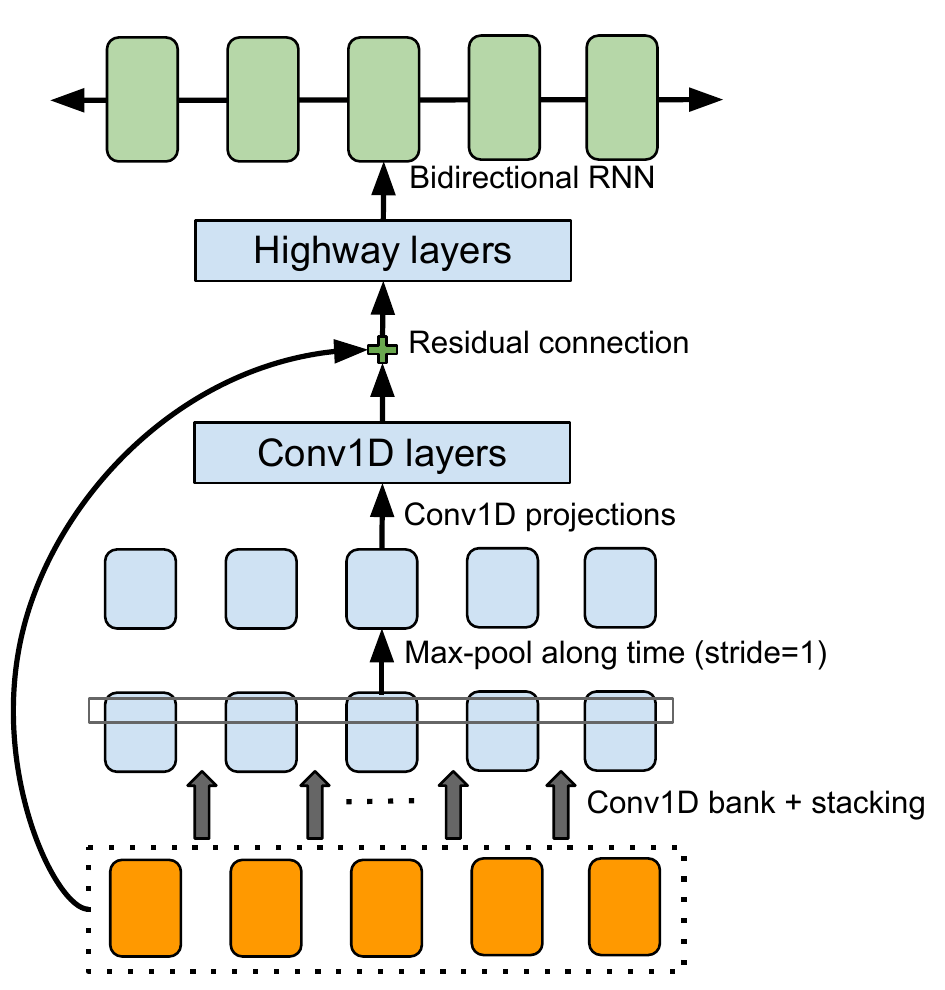}
\caption{The CBHG (1-D convolution bank + highway network + bidirectional GRU) module adapted from \cite{lee2016fully}.}
\label{fig:cbhg}
\end{figure}

\subsection{CBHG module}
We first describe a building block dubbed CBHG, illustrated in Figure \ref{fig:cbhg}. CBHG consists of a bank of 1-D convolutional filters, followed by highway networks \citep{srivastava2015highway} and a bidirectional  gated recurrent unit (GRU) \citep{chung2014empirical} recurrent neural net (RNN). CBHG is a powerful module for extracting representations from sequences. The input sequence is first convolved with $K$ sets of 1-D convolutional filters, where the $k$-th set contains $C_k$ filters of width $k$ (i.e. $k=1,2, \ldots, K$). These filters explicitly model local and contextual information (akin to modeling unigrams, bigrams, up to $K$-grams). The convolution outputs are stacked together and further max pooled along time to increase local invariances. Note that we use a stride of 1 to preserve the original time resolution. We further pass the processed sequence to a few fixed-width 1-D convolutions, whose outputs are added with the original input sequence via residual connections \citep{he2016deep}. Batch normalization \citep{ioffe2015batch} is used for all convolutional layers. The convolution outputs are fed into a multi-layer highway network to extract high-level features. Finally, we stack a bidirectional GRU RNN on top to extract sequential features from both forward and backward context. CBHG is inspired from work in machine translation \citep{lee2016fully}, where the main differences from \cite{lee2016fully} include using non-causal convolutions, batch normalization, residual connections, and stride=1 max pooling. We found that these modifications improved generalization. 

\subsection{Encoder}
The goal of the encoder is to extract robust sequential representations of text. The input to the encoder is a character sequence, where each character is represented as a one-hot vector and embedded into a continuous vector. We then apply a set of non-linear transformations, collectively called a ``pre-net", to each embedding. We use a bottleneck layer with dropout as the pre-net in this work, which helps convergence and improves generalization. A CBHG module transforms the pre-net outputs into the final encoder representation used by the attention module.  We found that this CBHG-based encoder not only reduces overfitting, but also makes fewer mispronunciations than a standard multi-layer RNN encoder (see our linked page of audio samples).

\subsection{Decoder}
We use a content-based tanh attention decoder (see e.g. \cite{vinyals2015grammar}), where a stateful recurrent layer produces the attention query at each decoder time step. We concatenate the context vector and the attention RNN cell output to form the input to the decoder RNNs. We use a stack of GRUs with vertical residual connections \citep{wu2016google} for the decoder. We found the residual connections speed up convergence. The decoder target is an important design choice. While we could directly predict raw spectrogram, it's a highly redundant representation for the purpose of learning alignment between speech signal and text (which is really the motivation of using seq2seq for this task). Because of this redundancy, we use a different target for seq2seq decoding and waveform synthesis. The seq2seq target can be highly compressed as long as it provides sufficient intelligibility and prosody information for an inversion process, which could be fixed or trained. We use 80-band mel-scale spectrogram as the target, though fewer bands or more concise targets such as cepstrum could be used. We use a post-processing network (discussed below) to convert from the seq2seq target to waveform. 

\begin{table}[t]
\centering
\caption{Hyper-parameters and network architectures. ``conv-$k$-$c$-ReLU" denotes
1-D convolution with width $k$ and $c$ output channels with ReLU activation. FC stands for fully-connected.}
\label{tb.params}
\scalebox{1.0}{
\begin{tabular}{l|l}
\hline
Spectral analysis & \emph{pre-emphasis}: 0.97; \emph{frame length}: 50 ms;\\
							& \emph{frame shift}: 12.5 ms; \emph{window type}: Hann \\ \hline
Character embedding & 256-D \\ \hline
Encoder CBHG & \emph{Conv1D bank}: $K$=16, conv-$k$-128-ReLU \\
						& \emph{Max pooling}: stride=1, width=2 \\
						& \emph{Conv1D projections}: conv-3-128-ReLU\\
						&  $\rightarrow$ conv-3-128-Linear \\
						& \emph{Highway net}: 4 layers of FC-128-ReLU \\
						& \emph{Bidirectional GRU}: 128 cells\\ \hline
Encoder pre-net & FC-256-ReLU $\rightarrow$ Dropout(0.5) $\rightarrow$ \\
						& FC-128-ReLU $\rightarrow$ Dropout(0.5) \\ \hline
Decoder pre-net & FC-256-ReLU $\rightarrow$ Dropout(0.5)$\rightarrow$ \\
						& FC-128-ReLU $\rightarrow$ Dropout(0.5) \\ \hline
Decoder RNN & 2-layer residual GRU (256 cells)\\ \hline
Attention RNN	& 1-layer GRU (256 cells) \\ \hline
Post-processing net  & \emph{Conv1D bank}: $K$=8, conv-k-128-ReLU \\
CBHG						& \emph{Max pooling}: stride=1, width=2 \\
						& \emph{Conv1D projections}: conv-3-256-ReLU\\
						&  $\rightarrow$ conv-3-80-Linear \\
						& \emph{Highway net}: 4 layers of FC-128-ReLU \\
						& \emph{Bidirectional GRU}: 128 cells\\ \hline
Reduction factor ($r$) & 2 \\ \hline
\end{tabular}
}
\end{table}

We use a simple fully-connected output layer to predict the decoder targets. An important trick we discovered was predicting multiple, non-overlapping output frames at each decoder step.  Predicting $r$ frames at once divides the total number of decoder steps by $r$, which reduces model size, training time, and inference time.  More importantly, we found this trick to substantially increase convergence speed, as measured by a much faster (and more stable) alignment learned from attention. This is likely because neighboring speech frames are correlated and each character usually corresponds to multiple frames. Emitting one frame at a time forces the model to attend to the same input token for multiple timesteps; emitting multiple frames allows the attention to move forward early in training. A similar trick is also used in \cite{zen2016fast} but mainly to speed up inference.

The first decoder step is conditioned on an all-zero frame, which represents a $<$GO$>$ frame. In inference, at decoder step $t$, the last frame of the $r$ predictions is fed as input to the decoder at step $t+1$.  Note that feeding the last prediction is an ad-hoc choice here -- we could use all $r$ predictions. During training, we always feed every $r$-th ground truth frame to the decoder. The input frame is passed to a pre-net as is done in the encoder. Since we do not use techniques such as scheduled sampling \citep{bengio2015scheduled} (we found it to hurt audio quality), the dropout in the pre-net is critical for the model to generalize, as it provides a noise source to resolve the multiple modalities in the output distribution.

\subsection{Post-processing net and waveform synthesis}
As mentioned above, the post-processing net's task is to convert the seq2seq target to a target that can be synthesized into waveforms. Since we use Griffin-Lim as the synthesizer, the post-processing net learns to predict spectral magnitude sampled on a linear-frequency scale. Another motivation of the post-processing net is that it can see the full decoded sequence. In contrast to seq2seq, which always runs from left to right, it has both forward and backward information to correct the prediction error for each individual frame. In this work, we use a CBHG module for the post-processing net, though a simpler architecture likely works as well. The concept of a post-processing network is highly general. It could be used to predict alternative targets such as vocoder parameters, or as a WaveNet-like neural vocoder \citep{van2016wavenet, mehri2016samplernn, arik2017deep} that synthesizes waveform samples directly.

We use the Griffin-Lim algorithm \citep{griffin1984signal} to synthesize waveform from the predicted spectrogram. We found that raising the predicted magnitudes by a power of 1.2 before feeding to Griffin-Lim reduces artifacts, likely due to its harmonic enhancement effect. We observed that Griffin-Lim converges after 50 iterations (in fact, about 30 iterations seems to be enough), which is reasonably fast. We implemented Griffin-Lim in TensorFlow \citep{abadi2016tensorflow} hence it's also part of the model. While Griffin-Lim is differentiable (it does not have trainable weights), we do not impose any loss on it in this work. We emphasize that our choice of Griffin-Lim is for simplicity; while it already yields strong results, developing a fast and high-quality trainable spectrogram to waveform inverter is ongoing work. 

\section{Model Details}
Table \ref{tb.params} lists the hyper-parameters and network architectures. We use log magnitude spectrogram with Hann windowing, 50 ms frame length, 12.5 ms frame shift, and 2048-point Fourier transform. We also found pre-emphasis (0.97) to be helpful. We use 24 kHz sampling rate for all experiments.

We use $r$ = 2 (output layer reduction factor) for the MOS results in this paper, though larger $r$ values (e.g. $r$ = 5) also work well. We use the Adam optimizer \citep{kingma2015adam} with learning rate decay, which starts from 0.001 and is reduced to 0.0005, 0.0003, and 0.0001 after 500K, 1M and 2M global steps, respectively. We use a simple $\ell$1 loss for both seq2seq decoder (mel-scale spectrogram) and post-processing net (linear-scale spectrogram). The two losses have equal weights.

We train using a batch size of 32, where all sequences are padded to a max length. It's a common practice to train sequence models with a loss mask, which masks loss on zero-padded frames. However, we found that models trained this way don't know when to stop emitting outputs, causing repeated sounds towards the end. One simple trick to get around this problem is to also reconstruct the zero-padded frames.

\section{Experiments}
We train Tacotron on an internal North American English dataset, which contains about 24.6 hours of speech data spoken by a professional female speaker. The phrases are text normalized, e.g. ``16" is converted to ``sixteen".

\begin{figure}[t]
\centering
\subfigure[Vanilla seq2seq + scheduled sampling]{
\label{fig:attn:vanilla}
\includegraphics[scale=0.6]{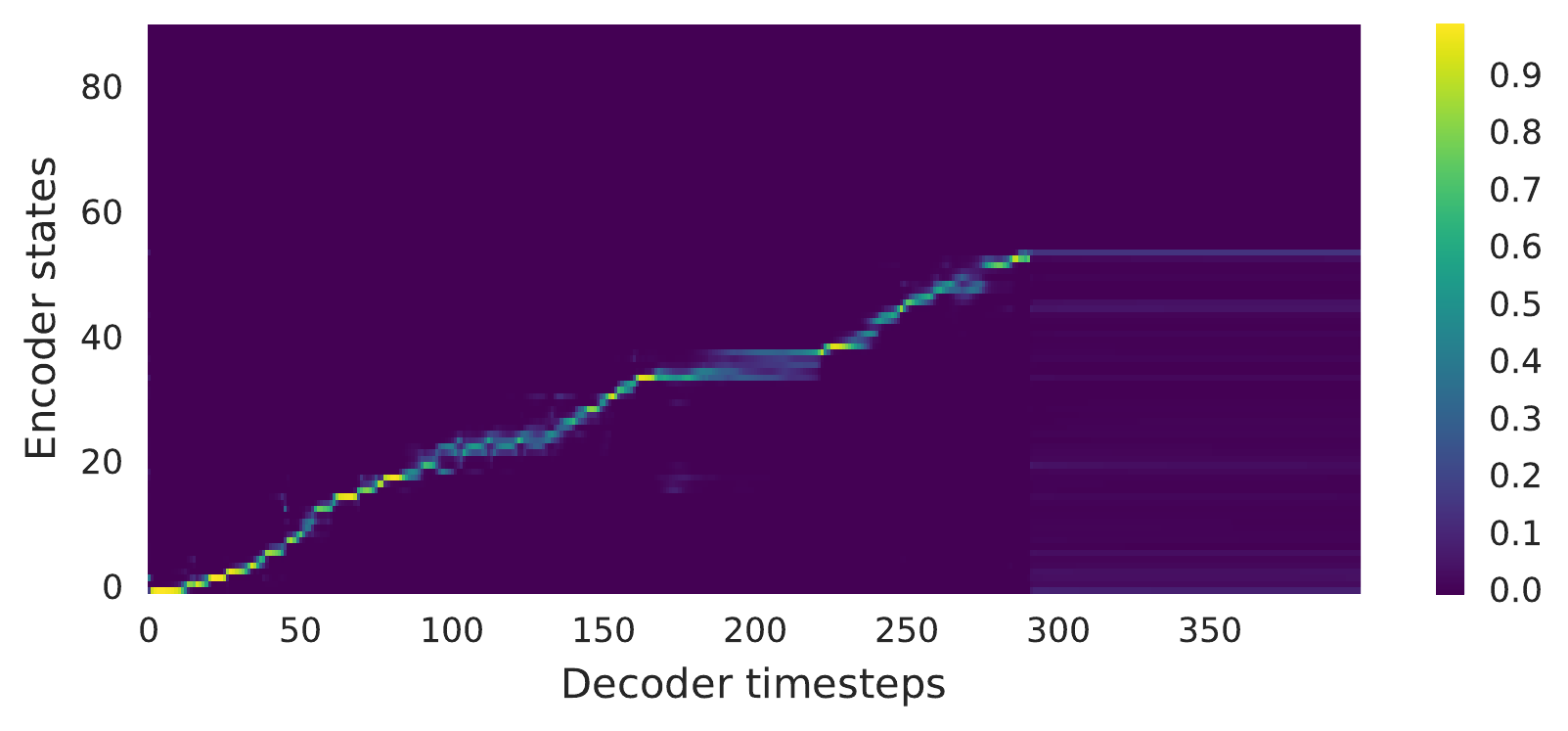}
}
\subfigure[GRU encoder]{
\label{fig:attn:gru}
\includegraphics[scale=0.6]{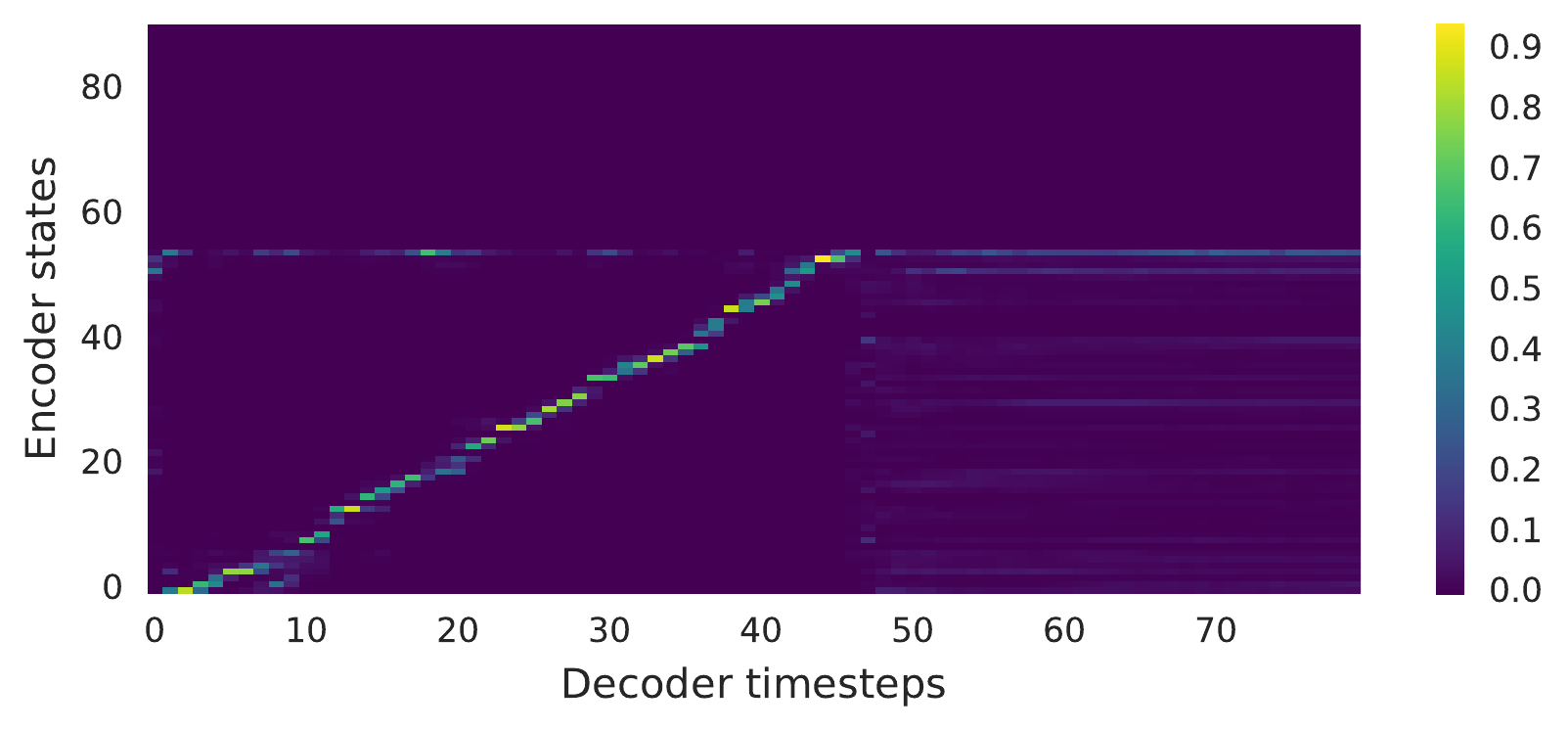}
}
\subfigure[Tacotron (proposed)]{
\label{fig:attn:proposed}
\includegraphics[scale=0.6]{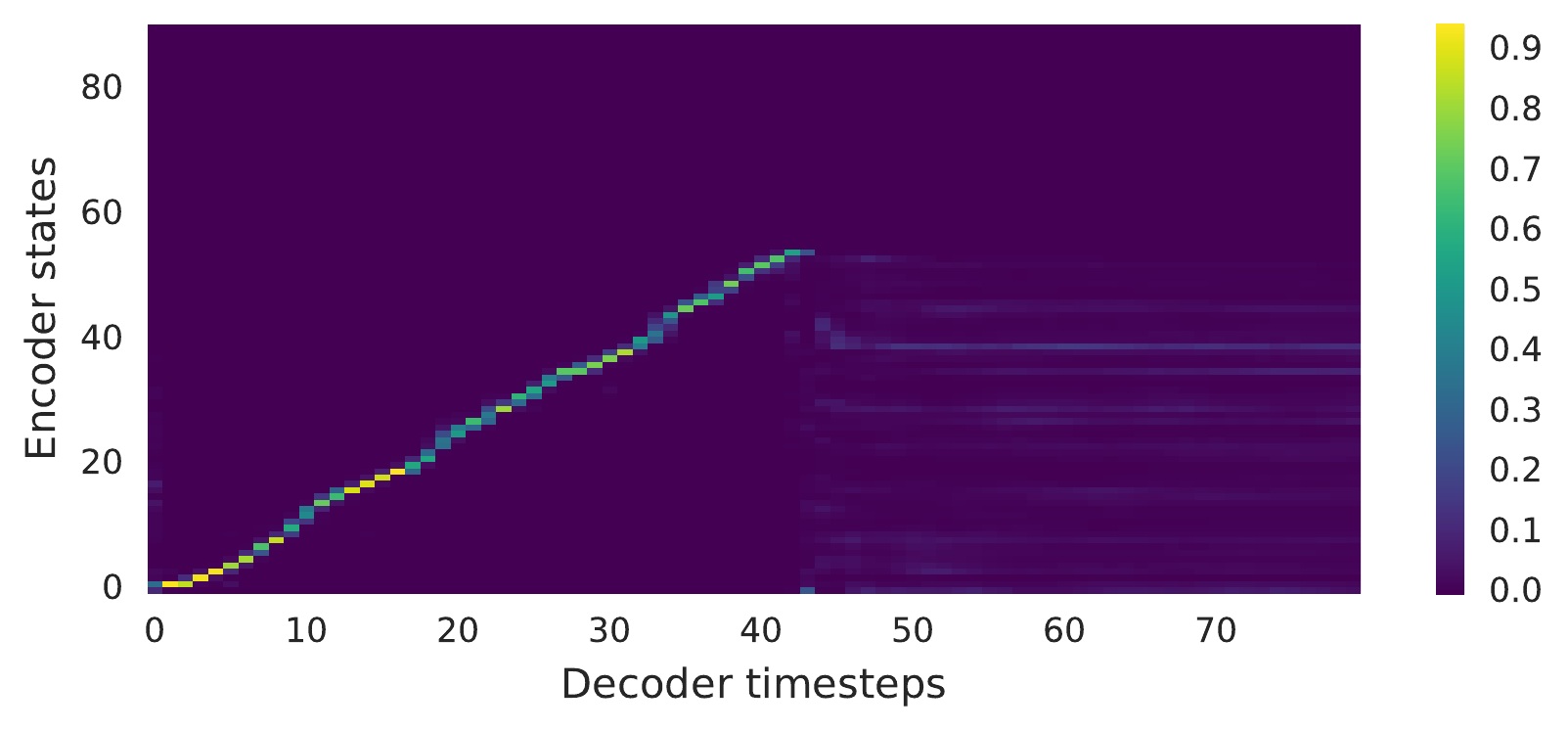}
}
\caption{{\it Attention alignments on a test phrase. The decoder length in Tacotron is shorter due to the use of the output reduction factor $r$=5. }}
\label{fig:attn}
\end{figure}

\subsection{Ablation analysis}
We conduct a few ablation studies to understand the key components in our model. As is common for generative models, it's hard to compare models based on objective metrics, which often do not correlate well with perception \citep{theis2015note}. We mainly rely on visual comparisons instead. We strongly encourage readers to listen to the provided samples.

First, we compare with a vanilla seq2seq model. Both the encoder and decoder use 2 layers of residual RNNs, where each layer has 256 GRU cells (we tried LSTM and got similar results). No pre-net or post-processing net is used, and the decoder directly predicts linear-scale log magnitude spectrogram. We found that scheduled sampling (sampling rate 0.5) is required for this model to learn alignments and generalize. We show the learned attention alignment in Figure \ref{fig:attn}. Figure \ref{fig:attn:vanilla} reveals that the vanilla seq2seq learns a poor alignment. One problem is that attention tends to get stuck for many frames before moving forward, which causes bad speech intelligibility in the synthesized signal. The naturalness and overall duration are destroyed as a result. In contrast, our model learns a clean and smooth alignment, as shown in Figure \ref{fig:attn:proposed}.

Second, we compare with a model with the CBHG encoder replaced by a 2-layer residual GRU encoder. The rest of the model, including the encoder pre-net, remain exactly the same. Comparing Figure \ref{fig:attn:gru} and \ref{fig:attn:proposed}, we can see that the alignment from the GRU encoder is noisier. Listening to synthesized signals, we found that noisy alignment often leads to mispronunciations. The CBHG encoder reduces overfitting and generalizes well to long and complex phrases.

\begin{figure}[t]
\centering
\subfigure[Without post-processing net]{
\label{fig:post:wo}
\includegraphics[scale=0.65]{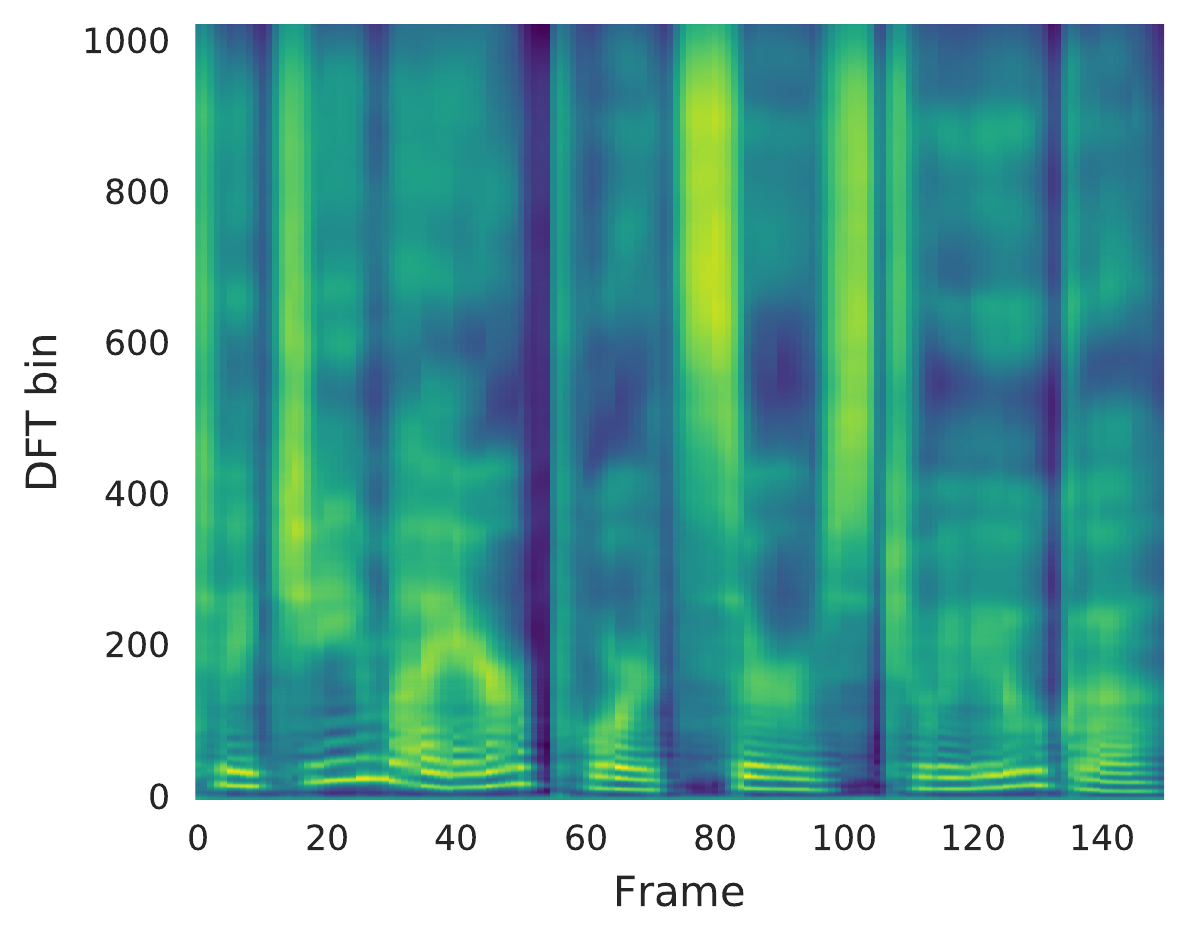}
}
\subfigure[With post-processing net]{
\label{fig:post:with}
\includegraphics[scale=0.65]{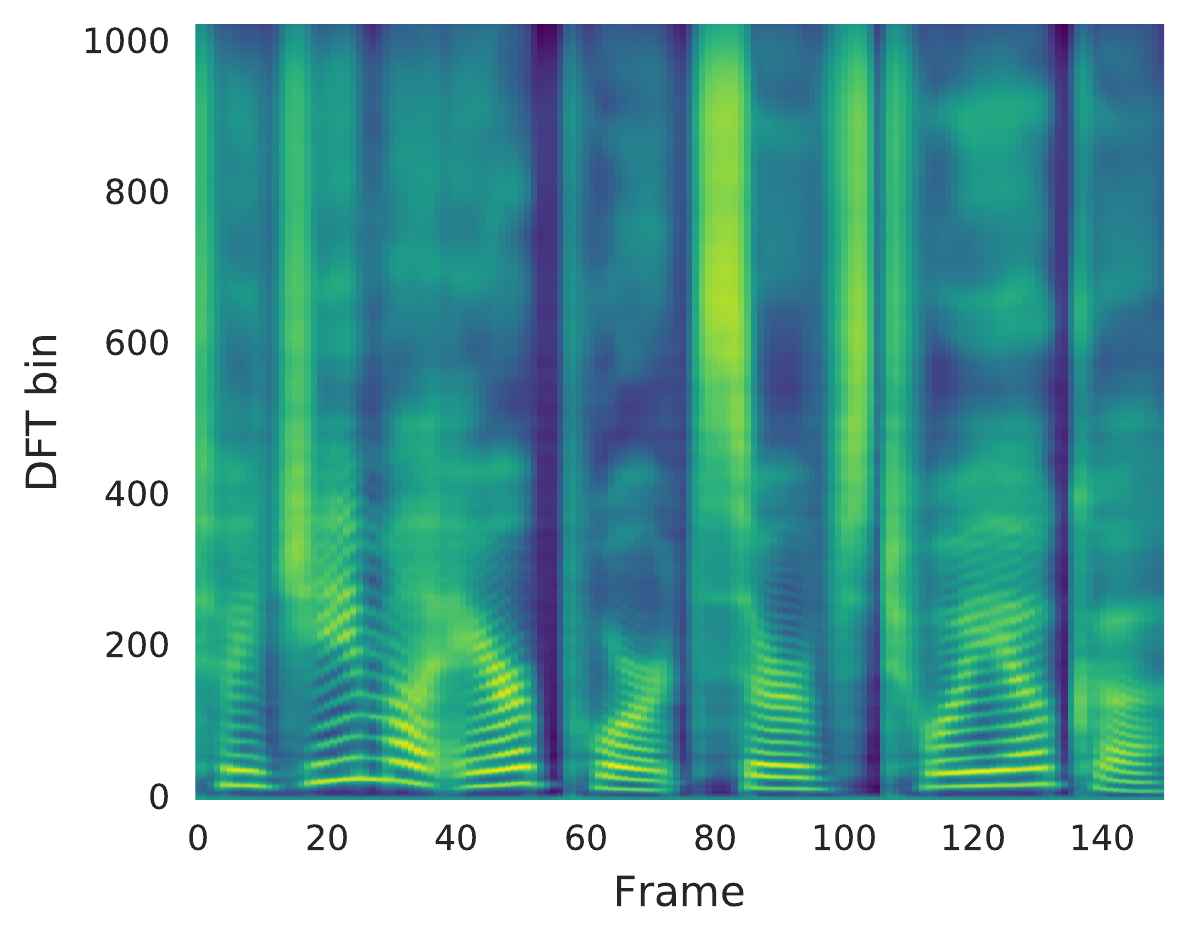}
}
\caption{{\it Predicted spectrograms with and without using the post-processing net.}}
\label{fig:post}
\end{figure}

Figures \ref{fig:post:wo} and \ref{fig:post:with} demonstrate the benefit of using the post-processing net. We trained a model without the post-processing net while keeping all the other components untouched (except that the decoder RNN predicts linear-scale spectrogram). With more contextual information, the prediction from the post-processing net contains better resolved harmonics (e.g. higher harmonics between bins 100 and 400) and high frequency formant structure, which reduces synthesis artifacts.

\subsection{Mean opinion score tests}
\label{sec.mos}
We conduct mean opinion score tests, where the subjects were asked to rate the naturalness of the stimuli in a 5-point Likert scale score. The MOS tests were crowdsourced from native speakers. 100 unseen phrases were used for the tests and each phrase received 8 ratings. When computing MOS, we only include ratings where headphones were used. We compare our model with a parametric (based on LSTM \citep{zen2016fast}) and a concatenative system \citep{gonzalvo2016recent}, both of which are in production. As shown in Table \ref{tb.mos}, Tacotron achieves an MOS of 3.82, which outperforms the parametric system. Given the strong baselines and the artifacts introduced by the Griffin-Lim synthesis, this represents a very promising result.

\begin{table}[htp!]
\centering
\caption{5-scale mean opinion score evaluation.}
\label{tb.mos}
\begin{tabular}{c|c}
\hline
 & mean opinion score \\ \hline
Tacotron & 3.82  $\pm$ 0.085  \\ \hline
Parametric & 3.69 $\pm$ 0.109 \\ \hline
Concatenative & 4.09  $\pm$ 0.119 \\ \hline
\end{tabular}
\end{table}
\section{Discussions}
We have proposed Tacotron, an integrated end-to-end generative TTS model that takes a character sequence as input and outputs the corresponding spectrogram. With a very simple waveform synthesis module, it achieves a 3.82 MOS score on US English, outperforming a production parametric system in terms of naturalness. Tacotron is frame-based, so the inference is substantially faster than sample-level autoregressive methods. Unlike previous work, Tacotron does not need hand-engineered linguistic features or complex components such as an HMM aligner. It can be trained from scratch with random initialization. We perform simple text normalization, though recent advancements in learned text normalization \citep{sproat2016rnn} may render this unnecessary in the future.

We have yet to investigate many aspects of our model; many early design decisions have gone unchanged. Our output layer, attention module, loss function, and Griffin-Lim-based waveform synthesizer are all ripe for improvement. For example, it's well known that Griffin-Lim outputs may have audible artifacts. We are currently working on fast and high-quality neural-network-based spectrogram inversion.

\subsubsection*{Acknowledgments}
The authors would like to thank Heiga Zen and Ziang Xie for constructive discussions and feedback.

\bibliography{winslow_bib}
\bibliographystyle{iclr2017_conference}

\end{document}